\documentclass{article}

\usepackage[preprint]{neurips_2020}

\usepackage[utf8]{inputenc}                                 
\usepackage[T1]{fontenc}                                    
\usepackage{xcolor}         
\definecolor{custom-blue}{RGB}{6,69,173} 
\usepackage[colorlinks, allcolors=custom-blue]{hyperref}
\usepackage{url}                                            
\usepackage{booktabs}                                       
\usepackage{amsfonts}                                       
\usepackage{amssymb}                                        
\usepackage{amsthm}                                         
\usepackage{mathtools}                                      
\usepackage{nicefrac}                                       
\usepackage{microtype}                                      
\usepackage{bm}                                             
\usepackage{lmodern}                                        %
\usepackage[font=small,labelfont=bf,textfont=it]{caption}   
\usepackage{subcaption}                                     
\usepackage[ruled,vlined]{algorithm2e}                      

\setcitestyle{authoryear,open={(},close={)}}

\theoremstyle{plain}
\newtheorem{definition}{Theorem}

\title{Deterministic Gaussian Averaged Neural Networks}

\usepackage{authblk}
\makeatletter
\renewcommand\AB@authnote[1]{\rlap{\textsuperscript{\normalfont#1}}}

\makeatother 

\author[*1]{\textbf{Ryan Campbell}\thanks{\textsuperscript{*} Correspondance to
\texttt{ryan.campbell2@mail.mcgill.ca}}}
\author[]{\textbf{Chris Finlay}}
\author[]{\textbf{Adam M Oberman}}
\affil[]{Department of Mathematics and Statistics, McGill University,
Montr\'eal, Canada}
\renewcommand\footnotemark{}

\begin{document}

\maketitle

\begin{abstract}
We present a deterministic method to compute the Gaussian average of neural networks used in regression and classification. Our method is based on an equivalence between training with a particular regularized loss, and the expected values of Gaussian averages. We use this equivalence to certify models which perform well on clean data but are not robust to adversarial perturbations. In terms of certified accuracy and adversarial robustness, our method is comparable to known stochastic methods such as randomized smoothing, but requires only a single model evaluation during inference. 
\end{abstract}

\section{Introduction}
Neural networks are very accurate on image classification tasks, but they are
vulnerable to adversarial perturbations, i.e. small changes to the model input leading to misclassification~\citep{szegedy2013intriguing}. 
Adversarial training \cite{madry2017towards} improves robustness, but does not solve the problem completely.  A complementary approach is \textit{certification}: find the greatest amount of perturbation that can be added to an input example before the model's performance is compromised \citep{lecuyer2019certified, raghunathan2018certified, cohen2019certified}. 
This means that, given a model $f$ and an input $x$, we wish to certify a minimum distance so that the classification of $f(x+\eta)$ is constant for all perturbations below the certified distance. 

One promising approach to certification is to certify
models within a given Gaussian region \citep{cohen2019certified, salman2019provably}. That is, a model $f$ should satisfy
\begin{equation}
	f(x) \approx \mathbb{E}_\eta \left[f(x+\eta)\right]
\end{equation} 
where $\eta\sim\mathcal{N}(0,\sigma^2I)$ and $x$ is a given example. While this is not easily attainable in practice, it is possible to obtain a new ``smoothed'' model $v$ that is the expected Gaussian average of our initial model $f$ at a given input example $x$,
\begin{equation}
	\label{eq:smoothed}
	v(x) \approx \mathbb{E}_\eta \left[f(x+\eta)\right]
\end{equation}
The works of \citeauthor{cohen2019certified} and \citeauthor{salman2019provably}
both obtain this Gaussian smoothed model \eqref{eq:smoothed} stochastically.
The idea is that by defining the classification of the stochastic model to be
the mode (most popular prediction) of the classifications given by ensembles
$f(x+\eta_1), f(x+\eta_2),\dots,$  we achieve a more robust ensemble.  While
ensemble models are generally more robust, the main advantage of the Gaussian ensemble is that we can obtain certified bounds based on probabilistic arguments.   However, like all ensemble models, it requires multiple inferences, which can be undesirable when computation is costly. 

In this work we present a \emph{deterministic} model whose outputs are
equivalent to the outputs of the Gaussian smoothed model $v(x)$ given by \eqref{eq:smoothed}.
  While it is generally not possible to capture an ensemble with a single model,
  in the special case of Gaussian smoothing, we can do so, due to specific
  properties of Gaussian averaging. Our deterministic smoothed
  model arises by training a regularized loss, which we call \textsc{HeatSmoothing}.    Our work relies on the insight, known since the early days of neural
networks \citep{bishop1995training, lecun1998LeNetpaper},  that gradient regularization is equivalent to Gaussian smoothing.  

Formally this is stated as follows.
\begin{definition}\label{thm:Bishop}
\citep{bishop1995training} Training a model using the quadratic loss, with added Gaussian noise of variance $\sigma^2$ to the inputs, is equivalent to training with 
\begin{equation}
	\label{Noise_Tychonoff}
	\mathbb E _x\left[\|f(x) - y\|^2
  + \sigma^2 \| \nabla f(x)\|^2  \right ] 
\end{equation}
up to higher order terms. 
\end{definition}
The proof is deferred to Appendix~\ref{app:proof-bishop}; a similar result is available with the cross-entropy loss.

\begin{figure}[t]
	\centering
  \includegraphics[width=.5\linewidth]{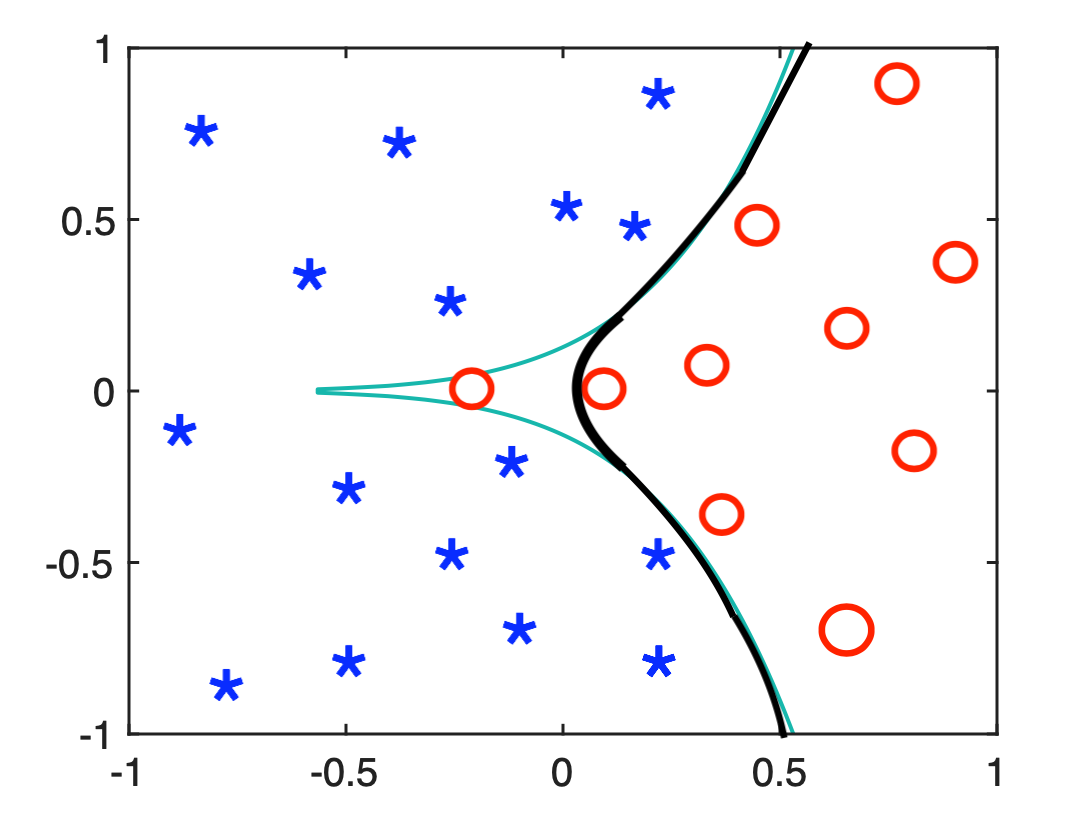} 
         \caption{Illustration of gradient regularization in the binary classification setting. The lighter line represents classification boundary for original model with large gradients, and the darker line represents classification boundary of the smoothed model.  The symbols indicate the classification by the original model: a single red circle is very close to many blue stars.  The smoothed model has a smoother classification boundary which flips the classification of the outlier.  }
        \label{fig:Cartoon}   
\end{figure}

The result from Theorem~\ref{thm:Bishop} gives an equivalence which is normally used to go from models augmented with Gaussian noise to regularized models.    In our case, we use the result in the other direction: we train a regularized model in order to produce a model which is equivalent to evaluating with noise. In practice, this means that rather than adding noise to regularize models for
certifiable robustness, we explicitly
perform a type of gradient regularization, \emph{in order to produce a model
which  performs as if Gaussian noise was added.} See \autoref{fig:Cartoon} for
an illustration of the effect of this gradient regularization.

To our knowledge, our method is the first deterministic Gaussian smoothing certification
technique. The main appeal of our approach is a large  decrease in the number of function
evaluations for computing certifiably robust models, and a
corresponding decrease in compute time at inference. Rather than stochastically sampling many times
from a Gaussian at inference time, our method is certifiably robust with only a
single model query.
Our improvement in inference time is achieved by iterative retraining of the model
so that model predictions are equal to the local Gaussian average of the model.
Although this iterative retraining procedure can be time consuming, we argue
that trading inference time for training time will be beneficial in many use case
scenarios. 


\section{Related work}


The issue of adversarial vulnerability arose in the works of
\citet{szegedy2013intriguing} and \citet{goodfellow2014explaining}, and has
spawned a vast body of research.
In terms of certification, early work by
\citet{cheng2017maximum} provided a method of
computing maximum perturbation bounds for neural networks, and reduced to
solving a mixed integer optimization problem. \cite{weng2018certification}
introduced non-trivial robustness bounds for fully connected networks, and provided tight robustness bounds at low computational cost. 
\cite{weng2018clever} proposed a metric that has theoretical grounding based on
Lipschitz continuity of the classifier model and is scaleable to state-of-the-art
ImageNet neural network classifiers. \cite{zhang2018crown} proposed a
general framework to certify neural networks based on linear and quadratic
bounding techniques on the activation functions, which is
more flexible than its predecessors.

    
Training a neural network with Gaussian noise has been shown to be equivalent to
gradient regularization \citep{bishop1995training}. This helps improve
robustness of models; however, there have been recent using noise for
certification purposes. \cite{lecuyer2019certified} first considered adding
random Gaussian noise as a certifiable defense in a method called
\textit{PixelDP}. In their method, they take a known neural network architecture
and add a layer of random noise to make the model's output random. The expected
classification is in turn more robust to adversarial perturbations. Furthermore,
their defense is a certified defense, meaning they provide a lower bound on the
amount of adversarial perturbations for which their defense will always work.
In a following work, \cite{li2018second} provided a defense with improved certified robustness. 
The certification guarantees given in these two papers are loose, meaning the
defended model will always be more robust than the certification bound indicates.

In contrast, \cite{cohen2019certified} provided a defense utilizing randomized Gaussian
smoothing that leads to \emph{tight} robustness guarantees under the $\ell_2$
norm. Moreover \citeauthor{cohen2019certified} used
Monte Carlo sampling to compute the radius in which a model's prediction is
unchanged; we refer to this method as \textsc{RandomizedSmoothing}.  
In work building on \citeauthor{cohen2019certified}, \cite{salman2019provably}
developed an adversarial training
framework called \textsc{SmoothAdv}  and defined a Lipschitz constant of averaged models. \cite{yang2020randomized} generalize previous randomized smoothing methods by providing robustness guarantees in the $\ell_1$, $\ell_2$, and $\ell_\infty$ norms for smoothing with several non-Gaussian distributions.


\begin{figure}[t]
	\centering
  	\begin{subfigure}{.33\textwidth}
        \centering
        \includegraphics[width=\linewidth]{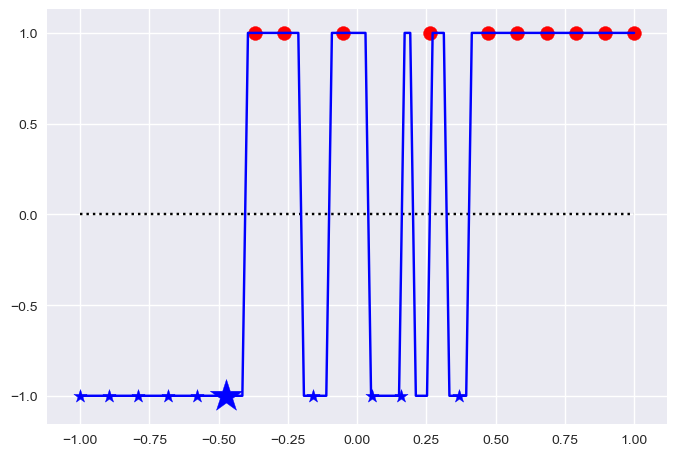}
        \caption{$f^0$}
        \label{fig:f0}
    \end{subfigure}%
    \begin{subfigure}{.32\textwidth}
        \centering
        \includegraphics[width=\linewidth]{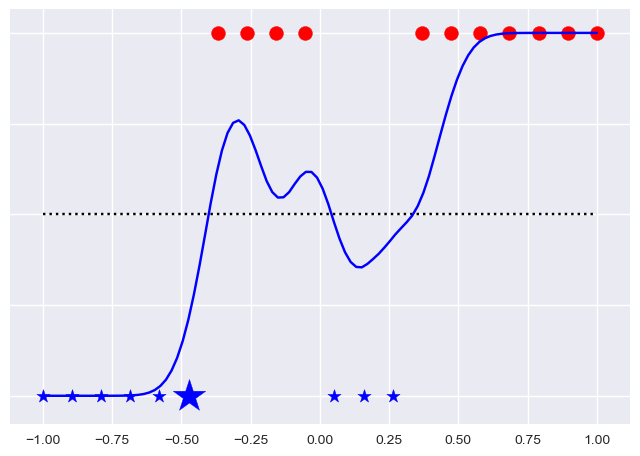}
        \caption{$f^1$}
        \label{fig:f1}
    \end{subfigure}
    \begin{subfigure}{.32\textwidth}
        \centering
        \includegraphics[width=\linewidth]{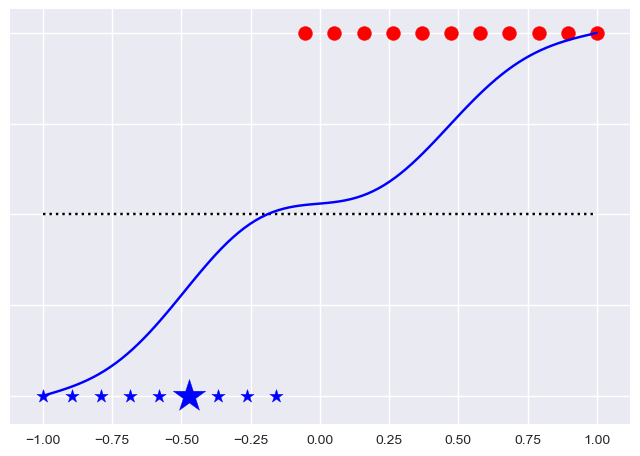}
        \caption{$f^5$}
        \label{fig:f1}
    \end{subfigure}
    \caption{Illustration of performing the iterative model update \eqref{eq:final-variational} for 5 timesteps in the binary classification setting. The dashed black line represents our decision boundary.   The blue line represents our current classification model. The blue stars and red circles represent our predicted classes using the current model iteration. Consider the datapoint at $x=-0.5$.  In the initial model $f^0$, the adversarial distance is $\approx0.10$. In model $f^5$, the adversarial distance is increased to $\approx0.35$.}
    \label{fig:heat-eqn-solv}   
\end{figure}

\section{Deterministic Smoothing}
Suppose we are given a dataset consisting of paired samples
$(x,y)\in\mathcal{X}\times\mathcal{Y}$ where $x$ is an example with corresponding true classification $y$. The supervised learning approach trains
a model $f:\mathcal{X}\longrightarrow\mathbb{R}^{Nc}$ which maps images to a
vector whose length equals the number of classes. Suppose  $f$  is the initial model,
and let $v$ be the averaged model given by equation \eqref{eq:smoothed}. \citet{cohen2019certified} find a Gaussian smoothed classification model $v$  by sampling $\eta\sim\mathcal{N}(0,\sigma^2I)$ independently $n$ times, performing $n$ classifications, and then computing the most popular classification. In the randomized smoothing method, the initial model $f$ is trained on data which is augmented with Gaussian noise to improve accuracy on noisy
images.

We take a different approach to Gaussian smoothing.
Starting from an accurate pretrained model $f$, we now discard the training
labels, and iteratively retrain a new model, $f^k$ using a quadratic loss
between the present iterate $f^k$ and the new model's predictions, with an
additional gradient regularization term.   
We have found that discarding the original one-hot
labels and instead using model predictions helps make the model smoother. 

To be precise, our iterative retraining procedure is given by
\begin{equation}
    \label{eq:final-variational}
    f^{k+1}=\underset{v}{\mathrm{argmin}} \;\mathbb{E}_x\left[ \frac{1}{2}\left\|v(x)-f^k(x)\right\|_2^2 + \lambda\frac{h\sigma^2}{2}\left\|\nabla_x v(x)\right\|_2^2\right]
\end{equation}
We set $f^0(x)=f(x)$, the initial model, and generate an iterative sequence of
retrained models for $k=0,\dots,n_{T}-1$, where $n_{T}$ is the number of
retraining steps. In \eqref{eq:final-variational}, $h={1}/{n_T}$ and $\lambda>0$ is a scaling hyperparameter. At each retraining step, the retrained model is progressively
smoothed by gradient regularization, while being encouraged to remain faithful
to the underlying base model via the quadratic loss. This is illustrated in Figure~\ref{fig:heat-eqn-solv}.

%
%


\subsection{Algorithmic details}
Note that the $\left\|\nabla_x v(x)\right\|_2^2$ term in
\eqref{eq:final-variational} requires the computation of a Jacobian matrix norm. In
high dimensions this is computationally expensive. To approximate this term, we
make use of the \textit{Johnson-Lindenstrauss lemma}
\citep{johnson1984extensions, vempala2005johnson-lindenstrauss} followed by the
finite difference approximation from \cite{finlay2019scaleable}. We are able to
approximate $\left\|\nabla_x v(x)\right\|_2^2$ by taking the average of the
product of the Jacobian matrix and Gaussian noise vectors. Jacobian-vector
products can be easily computed via reverse mode automatic differentiation, by
moving the noise vector $w$ inside: 
\begin{equation}
	\label{eq:moving-noise}
	w\cdot\left(\nabla_x v(x)\right)= \nabla_x \left(w\cdot v(x)\right)
\end{equation}
Further computation expense is reduced by using finite-differences to
approximate the norm of the gradient. Once the finite-difference is computed, we
detach this term from the automatic differentiation computation graph, further speeding training. More details of our implementation of these approximation techniques are presented in Appendix~\ref{app:jl}.

We have found that early on in training, the distance value
$\frac{1}{2}\left\|v(x)-f^k(x)\right\|_2^2$ may be far greater than the
$\frac{h\sigma^2}{2}\left\|\nabla_x v(x)\right\|_2^2$ term. We introduce a
scaling term $\lambda>0$ to correct this. We perform the training minimization
of \eqref{eq:final-variational} $n_T$ times for $k=0,\dots,n_T-1$, for a set
number epochs at each timestep $k$. The pseudo-code\footnote{Code is publicly
available at \url{https://github.com/ryancampbell514/HeatSmoothing}.} for our neural network weight update is given by Algorithm \ref{alg:heat-smoothing}.

\begin{algorithm}
 	\caption{\textsc{HeatSmoothing} Neural Network Weight Update}
    \label{alg:heat-smoothing}
    \SetKwInOut{Input}{Input}
    \SetKwInOut{Update}{Update}
    \SetKwInOut{Compute}{Compute}
    \SetKwInOut{Output}{Output}
	\SetKwInOut{Initialize}{Initialize}
	\SetKwInOut{Return}{Return}
	\SetKw{Continue}{continue}
	\SetKw{Break}{break}
    \SetAlgoLined
    \Input{Minibatch of input examples $\bm{x}^{(mb)}=\left(x^{(1)},\dots,x^{(Nb)}\right)$\\ 
    A model $v$ set to ``train'' mode \\ 
    Current model $f^{k}$ set to ``eval'' mode \\ 
    $\sigma$, standard deviation of Gaussian smoothing \\
    $\kappa$, number of Gaussian noise replications (default$=10$) \\
    $\delta$, finite difference step-size (default$=0.1$)\\
    $n_T$, total number of timesteps being executed (default$=5$)\\
    $\lambda\geq 1$, scaling hyperparameter}
    \Update{learning-rate according to a pre-defined scheduler.}
    \For{$i\in\left\{1,\dots Nb\right\}$}{
    \Compute{$v(x^{(i)}),f^{k}(x^{(i)})\in\mathbb{R}^{Nc}$\\
    		$J_i = \frac{1}{2}\left\|v(x^{(i)})-f^{k}(x^{(i)})\right\|_2^2\in\mathbb{R}$\\
    		\For{$j\in\left\{1,\dots\kappa\right\}$}{
    			Generate $w = \frac{1}{\sqrt{Nc}}\left(w_1,\dots,w_{Nc}\right)$, $w_1,\dots,w_{Nc}\in\mathcal{N}(0,1)$\\
    			Compute $l$ via \eqref{eq:l}, detach $x^{(i)}$ from the computation graph\\
				$J_i \leftarrow J_i + \lambda\frac{\sigma^2}{2n_T}\left(\frac{w\cdot v(x^{(i)}+\delta l) - w\cdot v(x^{(i)})}{\delta}\right)^2$    		
    		}
    		}
    		$J\leftarrow \frac{1}{Nb}\sum\limits_{i=1}^{Nb} J_i$}
    Update the weights of $v$ by running backpropagation on $J$ at the current learning rate.
\end{algorithm}

%

\subsection{Certified Radii}

To assess how well our model approximates the Gaussian average of the initial
model, we compute the certified $\ell_2$ radius introduced in
\cite{cohen2019certified}. A larger radius implies a better approximation of the
Gaussian average of the initial model. We compare our models with stochastically
averaged models via \textit{certified accuracy}. This is the fraction of the test set
which a model correctly classifies at a given radius while ignoring abstained
classifications. Throughout, we always use the same $\sigma$ value for certification as for
training. 

In conjunction with the certification technique of
\citeauthor{cohen2019certified}, we also provide the following theorem, which  describes a bound based on the Lipschitz constant of a Gaussian averaged model. We refer to this bound as the $L$-bound.

\begin{definition}[$\bm{L}$\textbf{-bound}]\label{thm:bound} 
  Suppose $v$ is the convolution (average) of $f:\mathbb{R}^d \to [0,1]^{Nc}$ with a Gaussian kernel of variance $\sigma^2 I$,
\[
v(x)=\left(f\ast\mathcal{N}(0,\sigma^2 I)\right)(x)
\] 
Then any perturbation $\delta$ which results in a change of rank of the $k$-th component of $v(x)$ must have norm bounded as follows: 
\begin{equation}
	\left\|\delta\right\|_2 \geq {\sigma}(\pi/2)^{1/2}  (v(x)_{(k)}-v(x)_{(k+1)})
\end{equation}
where $v(x)_{(i)}$ is the $i^\text{th}$ largest value in the vector $v(x)\in[0,1]^{Nc}$.
\end{definition} See Appendix~\ref{app:boundproof} for proof.
This bound is equally applicable to deterministic or stochastically averaged
models. 
In stochastically averaged models $v(x)$ is replaced by $\mathbb{E}_{\eta\sim\mathcal{N}(0,\sigma^2I)} \left[f(x+\eta)\right]$.

\subsection{Theoretical Details}
%

We appeal to partial differential equations (PDE) theory for explaining the
equivalence between gradient regularization and Gaussian convolution (averaging) of the
model\footnote{We sometimes interchange the terms Gaussian averaging and Gaussian
convolution; they are equivalent, as shown in Theorem
\ref{thm:heat-equiv}.}. The idea is that the gradient term which appears in the loss leads to a smoothing of the new function (model).  The fact that the exact form of the
smoothing corresponds to Gaussian convolution is a mathematical results which can be interpreted probabilistically or using techniques from analysis. Briefly, we detail the link as follows.

\cite{einstein1906theory} showed that the function value of an averaged model
under Brownian motion is related to
the heat equation (a PDE); the theory of stochastic differential equations makes this
rigorous \citep{karatzas1998brownian}. Moreover, solutions of the heat
equation are given by Gaussian convolution with the original model.
Crucially, in addition solutions of the
heat equation can be interpreted as iterations of a regularized loss problem
(called a variational energy) like that of Equation \ref{eq:final-variational}.  The
  minimizer of this variational energy
\eqref{eq:final-variational} satisfies an equation which is formally equivalent
to the heat equation \citep{gelfand2000calculus}. Thus, taking these facts
together, we see that a few steps of the minimization of the loss in
\eqref{eq:final-variational} yield a model which approximately satisfies the
heat equation, and corresponds to a model smoothed by Gaussian convolution. See
\autoref{fig:Cartoon} for an illustration of a few steps of the training
procedure. This result is summarized in the following theorem.
\begin{definition}\label{thm:heat-equiv} \citep{strauss2007partial} Let $f$ be a bounded function, $x\in\mathbb{R}^d$, and $\eta\sim\mathcal{N}\left(0,\sigma^2I\right)$. Then the following are equivalent:
    \begin{enumerate}
        \item $\mathbb{E}_\eta\left[f(x+\eta)\right]$, the expected value of Gaussian averages of $f$ at $x$.
        \item $\left(f\ast\mathcal{N}(0,\sigma^2I)\right)(x)$, the convolution of $f$ with the density of the $\mathcal{N}(0,\sigma^2I)$ distribution evaluated at $x$.
        \item The functions defined in 1. and 2. are solutions of the heat equation,
        \begin{equation}
            \label{eq:heat-eq}
            \frac{\partial}{\partial t}f(x,t) = \frac{\sigma^2}{2}\Delta_x f(x,t)
        \end{equation}
        at time $t=1$, with initial condition $f(x,0)=f(x)$.
    \end{enumerate}
\end{definition}

	 
In Appendix~\ref{app:proof-heat-thm}, we use Theorem~\ref{thm:heat-equiv} to show the equivalence of training with noise and iteratively training \eqref{eq:final-variational}.

\subsection{Adversarial Attacks}

To test how robust our model is to adversarial examples, we calculate the minimum $\ell_2$ adversarial via our $L$-bound and we attack our models using the \textit{projected gradient descent (PGD)} \citep{kurakin2016adversarial, madry2017towards} and \textit{decoupled direction and norm (DDN)} \citep{rony2018decoupling} methods. These attacks are chosen because there is a specific way they can be applied to stochastically averaged models \citep{salman2019provably}. In both attacks, it is standard to take the step
\begin{equation}
	\label{eq:pgd}
	g = \alpha \frac{\nabla_{\delta_t}L\left(f(x+\delta_t),y\right)}{\left\|\nabla_{\delta_t}L\left(f(x+\delta_t),y\right)\right\|_2}
\end{equation}
in the iterative algorithm. Here, $x$ is an input example with corresponding
true class $y$; $\delta_t$ denotes the adversarial perturbation at its current
iteration; $L$ denotes the cross-entropy Loss function (or KL Divergence);
$\varepsilon$ is the maximum perturbation allowed; and $\alpha$ is the step-size. In the stochastically averaged model setting, the step is given by
\begin{equation}
	\label{eq:pgd-stoch}
	g_n = \alpha \frac{\sum\limits_{i=1}^n\nabla_{\delta_t}L\left(f(x+\delta_t+\eta_i),y\right)}{\left\|\sum\limits_{i=1}^n \nabla_{\delta_t}L\left(f(x+\delta_t+\eta_i),y\right)\right\|_2}
\end{equation}
where $\eta_1,\dots,\eta_n\overset{\text{iid}}{\sim}\mathcal{N}(0,\sigma^2I)$. For our deterministically averaged models, we implement the update \eqref{eq:pgd}. This is because our models are deterministic, meaning there is no need to sample noise at evaluation time. For stochastically averaged models \citep{cohen2019certified, salman2019provably}, we implement the update \eqref{eq:pgd-stoch}.

\section{Experiments \& Results}

\subsection{CIFAR-10}

\begin{figure}
	\centering
    \begin{subfigure}{.5\textwidth}
        \centering
        \includegraphics[width=\linewidth]{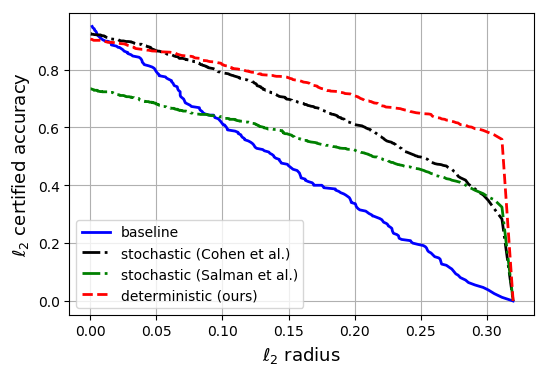}
        \caption{CIFAR-10 top-1 certified accuracy}
        \label{fig:cifar-cert}
    \end{subfigure}%
    \begin{subfigure}{.5\textwidth}
        \centering
        \includegraphics[width=\linewidth]{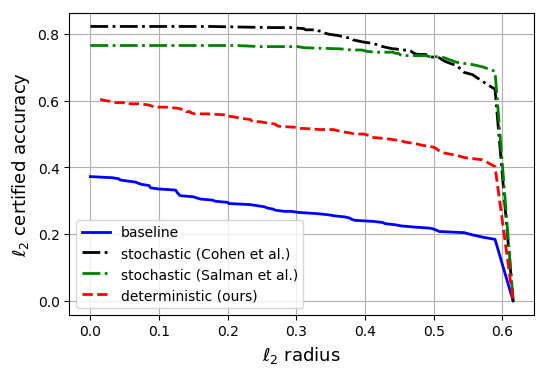}
        \caption{ImageNet-1k top-5 certified accuracy}
        \label{fig:imagenet-cert}
    \end{subfigure}
    \caption{Certified accuracy as a function of $\ell_2$ radius.}
    \label{fig:cert}
\end{figure}

We begin by testing our method on the CIFAR-10 dataset \citep{krizhevsky2009cifar10} with the ResNet-34 model architecture. The initial model $f$ was trained for 200 epochs with the cross-entropy loss function. Our smoothed model $v$ was computed by setting $f^0=f$ and running Algorithm~\ref{alg:heat-smoothing} with $\sigma=0.1$ and $\lambda=5$ for $n_T=5$ timesteps at 200 epochs each timestep. The training of our smoothed model took 5 times longer than the baseline model. We compare our results to a ResNet-34 model trained with $\sigma=0.1$ noisy examples as stochastically averaged model using \textsc{RandomizedSmoothing} \citep{cohen2019certified}. We also trained a \textsc{SmoothAdv} model \citep{salman2019provably} for 4 steps of PGD with the maximum perturbation set to $\varepsilon=0.5$. To assess certified accuracy, we run the \textsc{Certify} algorithm from \cite{cohen2019certified} with $n_0=100,n=10,000,\sigma=0.1$ for the stochastically trained models. For the \textsc{HeatSmoothing} model, we run the same certification algorithm, but without running \textsc{SamplingUnderNoise} to compute $\hat{c}_A$. For completeness, we also certify the baseline model $f^0$. Certification plots are presented in Figure~\ref{fig:cifar-cert}. In this plot, we see that our model's $\ell_2$ certified accuracy outperforms the stochastic models. Next, we attack our four models using PGD and DDN. We run both attacks with 20 steps and maximum perturbation $\varepsilon=4.0$ to force top-1 misclassification. Results are presented in Table~\ref{tab:cifar10-adv} and Figures~\ref{fig:cifar-pgd} and \ref{fig:cifar-ddn}. In Table~\ref{tab:cifar10-adv}, we see that \textsc{HeatSmoothing} outperforms the stochastic models in terms of robustness. The only exception is robustness to mean PGD perturbations. This is shown in Figures~\ref{fig:cifar-pgd}. Our model performs well up to an $\ell_2$ PGD perturbation of just above 1.0. 

\begin{table}
  \caption{$\ell_2$ adversarial distance metrics on CIFAR-10. A larger distance implies a more robust model.}
  \label{tab:cifar10-adv}
  \centering
  \begin{tabular}{lcccccc}
    \toprule
    Model        & \multicolumn{2}{c}{$L$-bound}   & \multicolumn{2}{c}{PGD}  & \multicolumn{2}{c}{DDN} \\
     & median & mean & median & mean & median & mean  \\
    \midrule
    \textsc{HeatSmoothing} & 0.094 & 0.085  & 0.7736 &0.9023 & 0.5358 &0.6361    \\
    \textsc{SmoothAdv}  & 0.090 & 0.078 & 0.7697 & 1.3241 & 0.4812 & 0.6208  \\
    \textsc{RandomizedSmoothing}  & 0.087 & 0.081 & 0.7425 &1.2677 & 0.4546 &0.5558  \\
    Undefended baseline & - & - & 0.7088 & 0.8390 & 0.4911 &0.5713 \\
    \bottomrule
  \end{tabular}
\end{table}

\begin{table}[b]
  \caption{$\ell_2$ adversarial distance metrics on ImageNet-1k}
  \label{tab:imagenet-adv}
  \centering
  \begin{tabular}{lcccccc}
    \toprule
    Model        & \multicolumn{2}{c}{$L$-bound}   & \multicolumn{2}{c}{PGD}  & \multicolumn{2}{c}{DDN} \\
     & median & mean & median & mean & median & mean \\
    \midrule
    \textsc{HeatSmoothing}  & 0.076 & 0.110 & 2.2857 & 2.3261 & 1.0313& 1.2102       \\
    \textsc{SmoothAdv}  & 0.160 & 0.160 & 3.5643 & 3.0244 & 1.1537 & 1.2850  \\
    \textsc{RandomizedSmoothing} & 0.200 & 0.180 & 2.6787 & 2.5587 & 1.2114 & 1.3412  \\
    Undefended baseline & - & - & 1.0313& 1.2832 & 0.8573& 0.9864   \\
    \bottomrule
  \end{tabular}
\end{table}

\subsection{ImageNet-1k}

We now execute our method on the ImageNet-1k dataset \citep{deng2009imagenet} with the ResNet-50 model architecture. The initial model $f$ was trained for 29 epochs with the cross-entropy loss function. Our smoothed model $v$ was computed by setting $f^0=f$ and running Algorithm~\ref{alg:heat-smoothing} with $\sigma=0.25$ and $\lambda=100$ for $n_T=5$ timesteps at 15 epochs each timestep. This took roughly 2 days per timestep. We note that the $ \frac{1}{2}\left\|v(x)-f^k(x)\right\|_2^2$ term is not suitable in the ImageNet-1k setting. This is due to the fact that the output vectors are too large, so the $\ell_2$ distance blows up early in training. To remedy this, we replace the $\ell_2$ distance with the Kullback-Leiblerd divergence between vectors $v(x)$ and $f^k(x)$.  Due to time constraints, we had to terminate training early; however, our model still attains a good top-5 accuracy. 

\begin{figure}[t]
	\centering
    \begin{subfigure}{.5\textwidth}
        \centering
        \includegraphics[width=\linewidth]{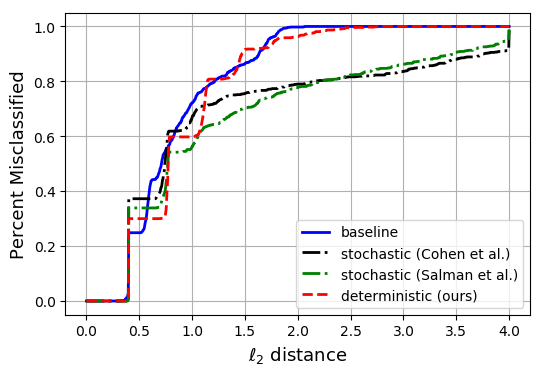}
        \caption{CIFAR-10 top-1 PGD}
        \label{fig:cifar-pgd}
    \end{subfigure}%
    \begin{subfigure}{.5\textwidth}
        \centering
        \includegraphics[width=\linewidth]{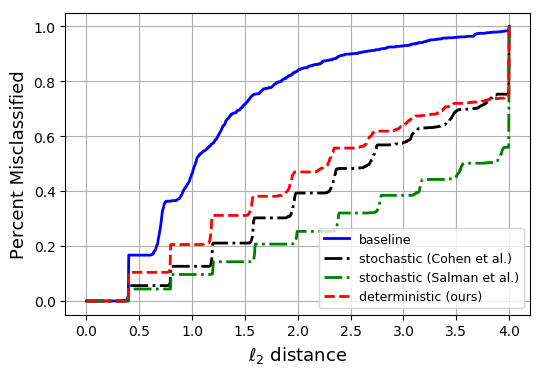}
        \caption{ImageNet-1k top-5 PGD}
    	\label{fig:imagenet-pgd}
    \end{subfigure}
    \begin{subfigure}{.5\textwidth}
        \centering
        \includegraphics[width=\linewidth]{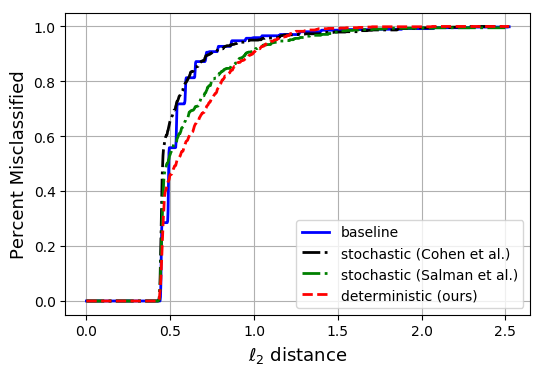}
        \caption{CIFAR-10 top-1 DDN}
    	\label{fig:cifar-ddn}
    \end{subfigure}%
    \begin{subfigure}{.5\textwidth}
        \centering
        \includegraphics[width=\linewidth]{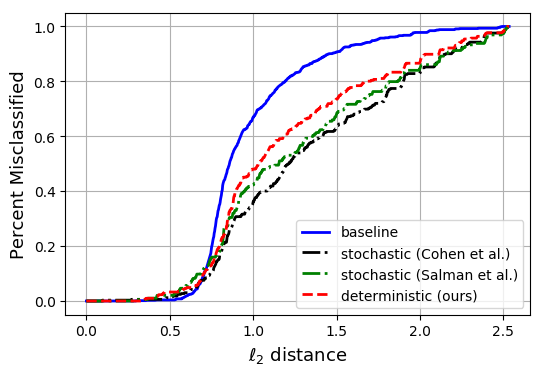}
        \caption{ImageNet-1k top-5 DDN}
    	\label{fig:imagenet-ddn}
    \end{subfigure}
    \caption{Attack curves: \% of images successfully attacked as a function of $\ell_2$ adversarial distance.}
    \label{fig:attack}
\end{figure}

We compare our results to a pretrained \textsc{RandomizedSmoothing} ResNet-50 model with $\sigma=0.25$ provided by \cite{cohen2019certified}. We also compare to a pretrained \textsc{SmoothAdv} ResNet-50 model trained with 1 step of PGD and with a maximum perturbation of $\varepsilon=0.5$ with $\sigma=0.25$ provided by \cite{salman2019provably}. To assess certified accuracy, we run the \textsc{Certify} algorithm from \cite{cohen2019certified} with $n_0=25,n=1,000,\sigma=0.25$ for the stochastically trained models. As in the CIFAR-10 setting, for the \textsc{HeatSmoothing} model, we run the same certification algorithm but without running \textsc{SamplingUnderNoise} to compute $\hat{c}_A$. For completeness, we also certify the baseline model $f^0$. Certification plots are presented in Figure~\ref{fig:imagenet-cert}. We see in this plot that due to the fact that we had to terminate our training early, our model has a lower certified accuracy compared to the pretrained stochastic models. Next, we attack our four models using PGD and DDN. We run both attacks until top-5 misclassification or until 20 steps are reached. Results are presented in Table~\ref{tab:imagenet-adv} and Figures~\ref{fig:imagenet-pgd} and \ref{fig:imagenet-ddn}. We see that our model is comparable to the stochastic models, but does not outperform them. We expect that with additional training iterations, we will be able to outperform the stochastic models.

\section{Conclusion}

Randomized smoothing is a well-known method to achieve a Gaussian average of
some initial neural network. This is desirable to guarantee that a model's
predictions are unchanged given perturbed input data. In this work, we used a
regularized loss to obtain deterministic Gaussian averaged models. By computing
$\ell_2$ certified radii, we showed that our method is comparable to
previously-known stochastic methods. By attacking our models, we showed that our
method is comparable to known stochastic methods in terms of adversarial
robustness of the resulting averaged model. We also developed a new lower bound
on perturbations necessary to throw off averaged models, and used it as a
measure of model robustness. Lastly, our method is less computationally
expensive in terms of inference time (see Appendix~\ref{app:time}).

Future work includes longer training time of our ImageNet-1k model. We are confident that once this is done, we will achieve comparable results to randomized smoothing methods on larger datasets. Secondly, adversarial training can be combined with our training method. This should result in deterministic averaged models that are also more robust to adversarial perturbations. Since our method is deterministic, we can adversarially train with any attack we choose without having to modify them to a stochastic averaging framework.

%


\small
\bibliography{bibliography}

\normalsize
\appendix

\newpage

\section{Proof of Theorem~\ref{thm:Bishop}}
\label{app:proof-bishop}
\begin{proof}
For clarity we treat the function as one dimensional. A similar calculation can be done in the higher dimensional case. Apply the Taylor expansion to the quadratic loss  $(f(x + v) - y)^2$, to give 
\[
f(x+\eta) = f(x) + \eta f_x + \frac 1 2 \eta ^2 f_{xx} + O(\eta^3)
\] 
Keeping only the lowest order terms, in $\eta$ and the lowest order derivatives, we obtain 
	\begin{align}
		(f(x + \eta) - y)^2 &= (f(x) - y)^2 + 2(f_x \eta  + \frac 1 2 \eta^2 f_{xx})(f(x)-y) + \text{ higher order terms }  
	\end{align}
Take expectation 
	\begin{align}
		\mathbb E \left[ (f(x + \eta) - y)^2  \right ] &= \mathbb E \left[(f(x) - y)^2 \right ] 
		 + \sigma^2 (f_x)^2  + \text{ higher order terms }  
	\end{align}
to drop the terms with odd powers of $v$ gives \eqref{Noise_Tychonoff}.
\end{proof}

\section{Solving the heat equation by training with a regularized loss function}
\label{app:proof-heat-thm}
Theorem~\ref{thm:heat-equiv} tells us that training a model with added Gaussian noise is equivalent to training a model to solve the heat equation. We can discretize the heat equation \eqref{eq:heat-eq} to obtain
\begin{equation}
    \label{eq:discreteheat}
    \frac{f^{k+1}-f^k}{h}=\frac{\sigma^2}{2}\Delta f^{k+1} 
\end{equation}
for $k=0,\dots,n_T-1$, where $n_T$ is the fixed number of timesteps, $h={1}/{n_T}$, and $f^0=f$, our initial model. Notice how, using the Euler-Lagrange equation, we can express $f^{k+1}$ in \eqref{eq:discreteheat} as the variational problem
\begin{equation}
    \label{eq:variational1}
    f^{k+1} = \underset{v}{\mathrm{argmin}} \;\frac{1}{2}\int\limits_{\mathbb{R}^d} \left(\left|v(x)-f^k(x)\right|^2 + \frac{h\sigma^2}{2}\left\|\nabla_x v(x)\right\|_2^2\right)\rho(x) dx 
\end{equation}
where $\rho$ is the density from which our clean data comes form. Therefore, this is equivalent to solving
\begin{equation}
    \label{eq:variational2}
    f^{k+1}=\underset{v}{\mathrm{argmin}} \;\mathbb{E}_x\left[ \left|v(x)-f^k(x)\right|^2 + \frac{h\sigma^2}{2}\left\|\nabla_x v(x)\right\|_2^2\right]
\end{equation}
Note that the minimizer of the objective of \eqref{eq:variational1} satisfies 
\begin{equation}
    \label{eq:min-variational}
    v-f^k = \frac{h\sigma^2}{2}\Delta v
\end{equation}
which matches \eqref{eq:discreteheat} if we set $f^{k+1}=v$. In the derivation of \eqref{eq:min-variational}, we take for granted the fact that empirically, $\rho$ is approximately uniform and is therefore constant. In the end, we iteratively compute \eqref{eq:variational2} and obtain models $f^1,\dots,f^{n_T}$, setting $v=f^{n_T}$, our smoothed model.

Something to take note of is that our model outputs be vectors whose length corresponds to the total number of classes; therefore, the objective function in \eqref{eq:variational2} will not be suitable for vector-valued outputs $f^k(x)$ and $v(x)$. We instead use the following update
\begin{equation}
    \label{eq:vector-variational}
    f^{k+1}=\underset{v}{\mathrm{argmin}} \;\mathbb{E}_x\left[ \frac{1}{2}\left\|v(x)-f^k(x)\right\|_2^2 + \frac{h\sigma^2}{2}\left\|\nabla_x v(x)\right\|_2^2\right]
\end{equation}

\section{Approximating the gradient-norm regularization term}
\label{app:jl}

By the Johnson-Lindenstrauss Lemma \citep{johnson1984extensions,vempala2005johnson-lindenstrauss}, $\left\|\nabla_x v(x)\right\|_2^2$ has the following approximation,
\begin{align}
	\begin{split}
        \left\|\nabla_x v(x)\right\|_2^2&\approx\sum\limits_{i=1}^{\kappa} \left\|\nabla_x \left(w_i\cdot v(x)\right)\right\|^2_2\\
        &\approx \sum\limits_{i=1}^{\kappa}\left(\frac{\left(w_i \cdot v\left(x+\delta l_i\right)\right)-\left(w_i\cdot v(x)\right)}{\delta}\right)^2
       \end{split}
\end{align}
where 
\begin{equation}
    \label{eq:w}
    w_i = \frac{1}{\sqrt{K}}\left(w_{i1},\dots,w_{iK}\right)^T\in\mathbb{R}^K\;,\;\;w_{ij}\overset{\text{iid}}{\sim}\mathcal{N}(0,1)
\end{equation}
and $l$ is given by
\begin{equation}
	\label{eq:l}
    l_i = \begin{cases}\frac{\nabla_x \left(w_i\cdot v(x)\right)}{\left\|\nabla_x \left(w_i\cdot v(x)\right)\right\|_2}&\text{if}\;\;\nabla_x \left(w_i\cdot v(x)\right)\neq 0\\0&\text{otherwise}\end{cases}
\end{equation}
In practice, we set $\delta=0.1$, $\kappa=10$, and $K=Nc$, the total number of classes. 

\section{Proof of Theorem~\ref{thm:bound}}
\label{app:boundproof}
\begin{proof}
Suppose the loss function $\ell$ is Lipschitz continuous with respect to model input $x$, with
Lipschitz constant $L$. Let $\ell_0$ be such that if $\ell(x)<\ell_0$, the model is always correct. Then by Proposition 2.2 in \cite{finlay2019scaleable}, a lower bound on the minimum magnitude of perturbation $\delta$ necessary to adversarially perturb an image $x$ is given by
\begin{equation}
	\label{eq:finlaybound}
	\left\|\delta\right\|_2\geq\frac{\max\left\{\ell_0 - \ell(x),0\right\}}{L}
\end{equation}
By Lemma 1 of Appendix A in \cite{salman2019provably}, our averaged model $$v(x)=\left(f\ast\mathcal{N}(0,\sigma^2I)\right)(x)$$ has Lipschitz constant $L=\frac{1}{\sigma}\sqrt{\frac{2}{\pi}}$. Replacing $L$ in \eqref{eq:finlaybound} with $\frac{1}{\sigma}\sqrt{\frac{2}{\pi}}$ and setting $\ell_0 = v(x)_{(k)}$, $\ell(x) = v(x)_{(k+1)}$ gives us the proof. \end{proof}

\section{Model inference computation time}
\label{app:time}

In Tables~\ref{tab:comp-time-CPU} and \ref{tab:comp-time-GPU}, we present the average time it takes to perform classification and certification of an image on the CPU and GPU, respectively. For certification, we implement the \textsc{Certify} algorithm from \cite{cohen2019certified}. On CIFAR-10, we use the top-1 classification setting while on ImageNet-1k, we use top-5 classification.

\begin{table}[h!]
  \caption{Average classification \& certification time on the CPU (seconds)}
  \label{tab:comp-time-CPU}
  \centering
  \begin{tabular}{lcccc}
    \toprule
    Model     &  \multicolumn{2}{c}{Classification}  & \multicolumn{2}{c}{Certification}  \\
     & CIFAR-10 & ImageNet-1k& CIFAR-10 & ImageNet-1k\\
    \midrule
    \textsc{HeatSmoothing} & 0.0049  & 0.0615 & 13.5994 & 3.2475  \\
    \textsc{RandomizedSmoothing}  & 0.0480  & 0.1631 &13.7394 &3.3938  \\
    \bottomrule
  \end{tabular}
\end{table}
\begin{table}[h!]
  \caption{Average classification \& certification time on the GPU (seconds)}
  \label{tab:comp-time-GPU}
  \centering
  \begin{tabular}{lcccc}
    \toprule
    Model     &  \multicolumn{2}{c}{Classification}  & \multicolumn{2}{c}{Certification}  \\
     & CIFAR-10 & ImageNet-1k& CIFAR-10 & ImageNet-1k\\
    \midrule
    \textsc{HeatSmoothing} & 0.0080  &  0.0113 & 0.3786  & 2.6624  \\
    \textsc{RandomizedSmoothing}  & 0.0399  & 0.0932 & 0.4142  & 2.8690 \\
    \bottomrule
  \end{tabular}
\end{table}

\end{document}